# From Group-Differences to Single-Subject Probability: Conformal Prediction-based Uncertainty Estimation for Brain-Age Modeling

J. Ernsting[1,2,3†], N. R. Winter[1], R. Leenings[1,3], K. Sarink[1], C. B. C. Barkhau[1], L. Fisch[1], D. Emden[1], V. Holstein[1], J. Repple[4,1], D. Grotegerd[1], S. Meinert[1,5], NAKO Investigators, K. Berger[6], B. Risse[2,3], U. Dannlowski[1], T. Hahn[1], for the Alzheimer's Disease Neuroimaging Initiative*

[1] University of Münster, Institute for Translational Psychiatry, Germany
[2] Institute for Geoinformatics, University of Münster, Germany
[3] Faculty of Mathematics and Computer Science, University of Münster, Germany
[4] Department for Psychiatry, Psychosomatic Medicine and Psychotherapy, University Hospital Frankfurt, Goethe University, Germany
[5] Institute for Translational Neuroscience, University of Münster, Germany
[6] Institute of Epidemiology and Social Medicine, University of Münster, Münster, Germany

*Data used in preparation of this article were obtained from the Alzheimer's Disease Neuroimaging Initiative (ADNI) database (adni.loni.usc.edu). As such, the investigators within the ADNI contributed to the design and implementation of ADNI and/or provided data but did not participate in analysis or writing of this report. A complete listing of ADNI investigators can be found at: http://adni.loni.usc.edu/wp-content/uploads/how_to_apply/ADNI_Acknowledgement_List.pdf

[†] Corresponding author:

Jan Ernsting, University of Münster, Institute for Translational Psychiatry, Germany

Albert-Schweitzer-Campus 1, D-48149 Münster

Phone: +49 (0)2 51 / 83 – 51859, E-Mail: j.ernsting@uni-muenster.de

## Abstract

The brain-age gap is one of the most investigated risk markers for brain changes across disorders. While the field is progressing towards large-scale models, recently incorporating uncertainty estimates, no model to date provides the single-subject risk assessment capability essential for clinical application. In order to enable the clinical use of brain-age as a biomarker, we here combine uncertainty-aware deep Neural Networks with conformal prediction theory. This approach provides statistical guarantees with respect to single-subject uncertainty estimates and allows for the calculation of an individual's probability for accelerated brain-aging. Building on this, we show empirically in a sample of N=16,794 participants that 1. a lower or comparable error as state-of-the-art, large-scale brain-age models, 2. the statistical guarantees regarding single-subject uncertainty estimation indeed hold for every participant, and 3. that the higher individual probabilities of accelerated brain-aging derived from our model are associated with Alzheimer's Disease, Bipolar Disorder and Major Depressive Disorder.

# Introduction

The rate at which aging-associated biological changes in the brain occur differs substantially between individuals. The so-called brain-age paradigm[1] aims to estimate a brain's biological age[2] and posits that accelerated brain age may serve as a cumulative marker of disease risk, functional capacity, and residual life span[3].

Based on a plethora of publications, this approach has developed into a cornerstone of biological age research linking the difference between chronological and brain-age to genetic, environmental, and demographic characteristics in health and disease [for a comprehensive review, see[4]]. For example, a higher *brain-age* compared to chronological age has been associated with markers of physiological aging (e.g., grip strength, lung function, walking speed), cognitive aging[5], life risk[6], and poor future health outcomes including progression from mild cognitive impairment to dementia[7,8], mortality[5], as well as a range of neurological diseases and psychiatric disorders[9].

In a typical brain-age study, a machine learning model is trained on neuroimaging data —usually whole-brain structural T1-weighted magnetic resonance imaging (MRI) data of a normative, most commonly healthy control group—to predict chronological age. This trained model is then used to evaluate neuroimaging data from previously unseen individuals and evaluated on the basis of the *brain-age gap* (BAG) as defined by the difference between predicted and chronological age.

With the BAG defined as the difference between predicted and chronological age, current studies are necessarily limited to group-level inference, e.g., identifying differences in mean BAGs between two groups such as healthy participants and patients. If, however, BAG should be employed as a risk-marker for a single subject – as is essential for any clinical use – individual probability of accelerated brain-aging needs to be quantified. In short, any clinical application of brain-age modeling requires a framework which enables the quantification of the probability with which a particular person shows accelerated brain-aging.

With current models, the deviation of an individual subject from a normative group could be calculated in theory, however, this deviation would not be meaningful as it might be confounded with training data density and availability. Additionally, higher deviation might be common in certain life spans (e.g. elderly or very young participants). Thereby a bias is introduced, which in current models cannot be controlled for.

In order to overcome the aforementioned shortcomings, a first step is to quantify single-subject uncertainty via a Monte Carlo dropout composite quantile regression (MCCQR) Neural Network architecture capable of estimating aleatory and epistemic uncertainty[10]. This approach takes into account confounding effects of epistemic and aleatory uncertainty.
While the resulting uncertainty estimates are empirically shown to capture underlying uncertainty well across numerous samples, they lack theoretical support and cannot provide statistical guarantees for a single person's confidence interval regarding their BAG. Extending the MCCQR approach, we therefore, present a novel framework combining an uncertainty-aware deep Neural Network model for brain age prediction with conformal prediction theory.

In short, conformal prediction theory is capable of providing statistically provable guarantees for the predicted uncertainty quantiles to cover the underlying true confidence interval. In other words, the Conformal Prediction algorithm is a statistically sound post-processing to further enhance predictions from an already fitted brain age model in terms of reliability. Based on this, our approach provides provable, statistical guarantees with respect to single-subject uncertainty estimates. Thus, it enables the calculation of individual probabilities for accelerated brain-aging with guaranteed uncertainty bounds.

Training on a normative population sample of N=10,691 participants from the German National Cohort (NAKO Gesundheitsstudie, NAKO)[11] and testing on four independent datasets of in total N=6,103 participants, we first test whether our model still performs comparably to existing state-of-the-art brain-age models. Second, we test empirically if the statistical guarantees regarding uncertainty estimation provided by the conformal prediction framework indeed hold for every participant. Finally, we quantify the confidence with which a person displays accelerated brain-aging and test whether a higher probability of accelerated brain-aging is associated with several pathological domains, including Major Depressive Disorder, Bipolar Disorder, mild cognitive impairment (MCI), and Alzheimer's disease (AD).

# Methods

## Training and validation samples

Whole-brain MRI data from five sources were used. We trained the MCCQR on the NAKO sample. Results for leave-site-out and 10-fold cross-validation are also based on the NAKO sample. Independent validation was based on the BiDirect sample[13], the Marburg-Münster Affective Disorders Cohort Study (MACS) data[15], the Information extraction from Images (IXI) dataset and the Alzheimer's Disease Neuroimaging Initiative (ADNI) data. In the following, we describe each dataset in more detail. Also, table S1 provides further sample characteristics, including sample sizes, gender distribution, age minimum and maximum, and SD.

## NAKO

NAKO is a large scale population study involving 205,000 study participants (age 20 to 72) from 18 study centers across Germany. The imaging program of the NAKO includes whole-brain coverage MRI (T1w-MPRAGE) in 30,000 participants recruited from 11 of the NAKO centers. This imaging program is performed in five NAKO imaging centers equipped with identical MR scanners (3.0 T Skyra, Siemens Healthineers, Erlangen, Germany). The brain MRI data sets used in this work were derived from the "data freeze 100K" milestone of the first 100,000 participants. This cohort includes the first 10,691 NAKO participants with completed MRIs of proven image quality [for a detailed protocol, see [11,12]]. BMI was calculated from the participants height and weight (kg/m²; mean = 26.82; SD = 4.76).

## BiDirect

The BiDirect study is an ongoing study that comprises three distinct cohorts: patients hospitalized for an acute episode of major depression, patients two to four months after an acute cardiac event, and healthy controls randomly drawn from the population register of the city of Münster, Germany. Baseline examination of all participants included a structural MRI of the brain (Gyroscan Intera 3T, Philips Medical Systems, the Netherlands), a computer-assisted face-to-face interview about sociodemographic characteristics, a medical history, an extensive psychiatric assessment, and collection of blood samples. Inclusion criteria for the present study were availability of completed baseline MRI data with sufficient MRI quality. All patients with major depressive disorder had an episode of major depression at the time of recruitment and were either currently hospitalized (>90%) or had been hospitalized for depression at least once during the 12 months before inclusion in the study (<10%). Further details on the rationale, design, and recruitment procedures of the BiDirect study have been described here [13].

## Marburg-Münster Affective Disorder Cohort Study

Participants were recruited through inpatient and outpatient services of local psychiatric hospitals or newspaper advertisements. All patients included here suffered from a major depressive disorder or bipolar disorder according to DSM-IV criteria, with the current episode being acute or (partially) remitted. Healthy controls were free from any life-time psychiatric diagnosis. Participants with any history of neurological (e.g., concussion, stroke, tumor, neuro-inflammatory diseases) or medical (e.g., cancer, chronic inflammatory or autoimmune diseases, heart diseases, diabetes mellitus, infections) conditions were excluded. MRI data was acquired using two 3T whole body MRI scanner (Marburg: MAGNETOM Trio Tim, software version Syngo MR B17, Siemens, Erlangen, Germany, 12-channel head matrix Rx-coil; Münster: MAGNETOM Prisma, software version Syngo MR D13D, Siemens, Erlangen, Germany, 20-channel head matrix Rx-coil). Further details about the sample, structure, and assessment protocols of the MACS[14] and MRI quality assurance protocol[15] are provided elsewhere. Inclusion criteria for the present study were availability of completed baseline MRI data with sufficient MRI quality [see [15] for details].

## Information eXtraction from Images

This dataset comprises MRI data from healthy participants, along with demographic characteristics, collected as part of the IXI project available for download (https://brain-development.org/ixi-dataset/). The data have been collected at three hospitals in London (Hammersmith Hospital using a Philips 3T system, Guy's Hospital using a Philips 1.5T system, and Institute of Psychiatry using a GE 1.5T system). Inclusion criteria for the present study were availability of completed baseline MRI data.

## Alzheimer's Disease Neuroimaging Initiative

The ADNI dataset consists of data composed of four cohorts: ADNI 1, ADNI 2, ADNI Go and ADNI 3. Since data collection for ADNI 3 has not ended yet, our dataset contained all four cohorts and images in ADNI 3 available on February 2nd, 2022. ADNI contains images from healthy controls and patients suffering from Mild Cognitive Imparment (MCI) and Altzheimer's disease. The ADNI data have been curated and converted to the Brain Imaging Data Structure (BIDS) format[16] using Clinica[17,18].

## MRI preprocessing

MRI data were preprocessed using the CAT12 toolbox (built 1450 with SPM12 version 7487 and Matlab 2019a; http://dbm.neuro.uni-jena.de/cat) with default parameters. Images were bias corrected, segmented using tissue classification, normalized to MNI-space using DARTEL normalization, smoothed with an isotropic Gaussian Kernel (8 mm FWHM), and resampled to 3-mm isomorphic voxels. Using the PHOTON AI software[19] (see "Model training and validation" section below), a whole-brain mask comprising all gray matter voxels was applied, data were vectorized, features with zero variance in the NAKO dataset were removed, and the scikit-learn Standard Scaler was applied.

## MCCQR model

The Monte Carlo dropout composite quantile regression (MCCQR) model was initially introduced in [10] and has been used as a baseline model for addition of the conformal prediction technique. In contrast to most machine learning algorithms, the MCCQR is able to provide uncertainty estimates for high dimensional MRI data, whereas current state-of-the-art methods like Gaussian Progress Regression or Relevance Vector Regression fail to provide those. The model combines regression quantiles for aleatory uncertainty estimation with Monte Carlo Dropout for epistemic uncertainty estimation.

## Conformal Predictions

Whilst the MCCQR model provides uncertainty estimates, as outlined above, predicted quantiles might be inaccurate. To account for that we used the conformal predictions framework as presented by Vovk et al.[20,21] described in [22]. The conformal predictions framework provides statistical proof that the underlying confidence interval is actually covered by the quantiles. This requires an additional hold-out set, called calibration set, which consists of i.i.d. data. Given the calibration set we can create a candidate prediction set (i.e., quantiles for regression tasks) which then holds the guarantee to capture the true quantile values with probability

$$1 - \alpha < \mathbb{P}(Y_{test} \in C(X_{test})) \leq 1 - \alpha + \frac{1}{1 + n}$$

given a novel, unseen data point $(X_{test}, Y_{test})$.

Given the calibration set we are able to calculate a score function for all calibration data points

$$s(x, y) = max(\hat{y}_{\alpha/2}(x) - y, y - \hat{y}_{1-\alpha/2}(x)).$$

Once we calculated the scores on the calibration set, we can calibrate the model by computing $\hat{q}$, a calibration constant for a single quantile by simply taking

$$\hat{q} = Quantile[s_1, \ldots, s_n; \frac{(n+1)(1-\alpha)}{n}].$$

Using the calculated calibration constant, we can then calibrate the models predictions by adding or subtracting the constant on the respective quantile

$$C(x) = [\hat{y}_{\alpha/2}(x) - \hat{q}, \hat{y}_{1-\alpha/2}(x) + \hat{q}].$$

As the MCCQR model is capable of providing multiple quantile estimations (101), we simply calibrate all available quantiles with their respective error rates (e.g. the 90% confidence interval between the 5% and 95% quantile should actually contain 90% of the real data points resulting in $\alpha = 10$).

## Machine Learning model

Similar to the MCCQR model we trained a model consisting of 32 rectified linear units using Tensorflow 2.0 and tensorflow probability. We trained for 10 epochs with a learning rate of 0.01, a batch size of 64, and dropout rate of 0.2 using the Adam Optimizer with default settings. In order to acquire expressive quantile coverage we choose calibration set size n=1000, resulting in $1 - \alpha < \mathbb{P}(Y_{test} \in C(X_{test})) \leq 1 - \alpha + 0.0009$ and thereby no unintended crossover of quantile boundaries. For estimation of epistemic and aleatory uncertainty we kept sampling 1000 times for each sample. The final prediction was computed as the mean on the quantiles after application of conformal predictions refinement. The procedure of model training and confirmization is shown in Figure 1.

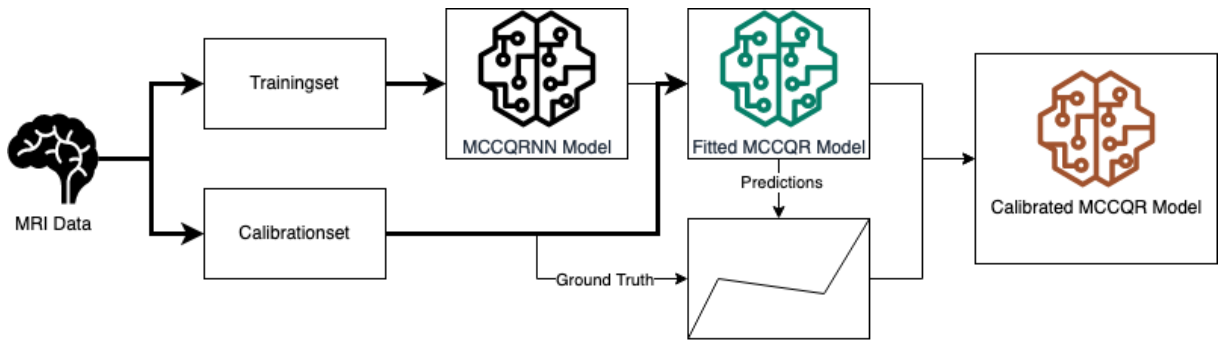

Figure 1: Overview of the machine learning model. The input data (NAKO dataset) is splitted into two different sets: training- and calibration-set. The model is fitted on the trainingset as usually and the predictions for the calibration set are then used to construct the network output leading to a finally calibrated model.

# Results

## Model Performance

First, we evaluated our model against four independent datasets covering a broader age range than the training dataset. We observed a final MAD of 2.92 (0.11) in the Cross Validation and MAD of 2.94 (0.28) in the leave-site-out cross validation. As our model is based on the MCCQR Model presented in [10], we will use this model as a baseline. Note that the MCCQR model outperformed all existing models as shown in [10]. We show that our model outperforms the MCCQR in four of five tasks. In the external datasets our model also reaches state-of-the-art performance: In IXI our model reaches a MAD of 4.40, in the MDD sample of the MACS a MAD of 3.71 and in the BiDirect our model obtains a MAD of 3.24. For a comprehensive analysis of baseline architectures and classic brain-age model performances on the same datasets please see[10].

| Model | 10 Fold CV | Leave-site-out CV | MACS HC only | BiDirect | IXI |
|---|---|---|---|---|---|
| MCCQR | 2.95 (0.16) | **2.94 (0.22)** | 3.91 | 3.45 | 4.57 |
| Conformal | **2.92 (0.11)** | 2.94 (0.28) | **3.69** | **3.02** | **4.40** |

Table1: MCCQR baseline and Conformal Predictions model results. Best results for each dataset are bold.

## Uncertainty Quantification

To evaluate the model's performance with regard to uncertainty quantification, we calculated the prediction interval coverage probabilities (PICPs), i.e., the probability that the predicted quantile contains the true value of the predicted sample. To calibrate the final predictions for uncertainty, we crucially need accurate uncertainty estimates from our model. As the conformal predictions framework is forcing our model to provide highly accurate uncertainty estimates, we expect the model to provide more conservative uncertainty quantiles. This behavior can be observed in Supplementary Figure 1. As expected, compared to the original MCCQR estimates, the model's predictions are much more conservative after applying the conformal prediction algorithm. Whilst being more conservative, the predictions are still meaningful, i.e., not covering 100% in all quantiles providing a trivial solution to enforcing quantiles covering the true underlying confidence interval. Most importantly, all subjects are included in their respective expected interval.

## Individual probability of accelerated brain aging

With the exception of [10] and [23], the current brain-age literature only use the point estimate of the brain age prediction to obtain a brain age gap which is then associated with various risk markers. Whilst appealing in theory, application on individual subject level is impossible when using this approach. We aimed to estimate the subject wise probability of accelerated brain aging by calculating the $\alpha$-level of the person, i.e. the first quantile containing the true age of the subject (for details on $\alpha$-level calculation see[22]). By assigning a probability of accelerated brain aging to each quantile we obtained a probability score in [-50, 50], encoding how probable accelerated brain aging is in the current subject. Negative values are assigned for negative brain age deviation.

For the analysis of group differences we hypothesized that a higher probability value should be an indicator for a major depressive disorder or bipolar disorder compared to healthy controls. We were then able to show a significant difference between the probability value of patients suffering from major depressive disorder and healthy controls from the MACS dataset (U=355330.5, p=0.020) and a significant difference between the probability values of patients suffering from bipolar disorder and healthy controls (U=47357.5, p<0.001) using a Mann-Whitney rank test, with healthy subject showing lower probability for accelerated brain aging.

Finally we used the ADNI dataset to test our second hypothesis, that a higher probability of brain aging is an indicator for MCI or AD. Using a Man-Whitney rank test we were able to show a significant difference between MCI patients and healthy controls from ADNI (U=331487.5, p<0.001), a significant difference between AD patients and MCI subjects (U=212510, p<0.0469) and a significant difference between healthy controls and AD patients (U=194395.5, p<0.001), with healthy controls showing lower probability for accelerated brain aging.

## Discussion

This work presents a novel framework combining an uncertainty aware Neural Network model for brain age prediction with conformal prediction theory, which provides statistical guarantees with regard to single-subject uncertainty estimates. To this end, we trained a version of the MCCQR model on a normative population sample of N=10,691 participants from the NAKO and tested it on four independent datasets of in total N=6,103 participants. First, we show that our model outperforms existing brain-age models providing a lower or comparable error. Second, we show that the statistical guarantees regarding uncertainty estimation obtained by combining MCCQR with conformal prediction theory, indeed hold for every participant (see Figure S1). Finally, we quantify the confidence with which a person displays accelerated brain-aging and show that higher probabilities of accelerated brain-aging are associated with several pathological domains. We find an association with mental health diagnoses for Major Depressive Disorder as well as Bipolar Disorder. Additionally, we show that higher probabilities of accelerated brain-aging are associated with mild cognitive impairment (MCI) and Alzheimer's disease (AD).

This approach enables the calculation of an individual's probability for accelerated brain-aging, thereby allowing for the quantification of individual risk. This, in turn, enables the use of brain-age estimation as a risk mark in clinical practice for the first time.

Methodically, the current approach is an extension of the MCCQR model[10]. The MCCQR model is capable of providing uncertainty estimates based on the estimation of aleatory and epistemic uncertainty combined, using a single loss function for multiple quantiles at once. This allows correction for training data density and availability effects. Correction for uncertainty is an important step, as uncertainty might be confounding the predictions leading to higher BAG caused by characteristics of the training set and properties of the model. This constitutes a challenge in clinical applications, where the model's predictions are intended to provide diagnostic or prognostic insight about patients or subjects. These clinical use cases and potential insights for individual subject screening have to be identified in future studies.

Uncertainty estimation is a powerful tool to provide insight into the model and into the underlying training data density and availability effects. However, uncertainty estimates for models such as the MCCQR can only be validated empirically and thus lack methodical rigor in a statistical sense with regard to the confidence intervals. While the MCCQR Model provides uncertainty estimates which are shown to be valid on the NAKO dataset, the model's predictions are not guaranteed to provide valid uncertainty estimations. Combining the MCCQR approach with conformal prediction theory, however, provides guarantees leading to a methodologically more robust model in practice.

In addition, predictions based on our approach can be interpreted as the individual probability of a sample showing accelerated brain age. While, in theory, this is also possible using the MCCQR model, the MCCQR based predictions are not mathematically proven to capture the true underlying confidence intervals. This renders the model's predictions potentially unreliable when aiming to provide a novel clinical usable biomarker. The novel model based on conformal predictions is now able to provide statistical guarantees leading to a trustworthy but more conservative model.

More generally, brain-age estimation can be seen as an application of normative modeling. Normative modeling utilizes a representative sample of neurotypical or healthy control subjects against which individual subjects are then compared. First a model is fitted to the data of the healthy control groups data. The expectation is, that the model is capable of learning the characteristics of healthy or normal aging. In a second step this model is then shown the data of subjects with different characteristics, most commonly subjects with diagnoses. If the diagnosed disease causes brain changes, we expect the model to predict an age that is higher or lower than the biological age of the currently observed person. Thus, normative modeling requires normative samples to provide meaningful predictions.

Here, we use the NAKO data as a population-based cohort upon which we build our model. As participants are largely representative of the population in Germany, prevalence for common diseases and risk factors may be comparable with existing epidemiological studies. Thus, the resulting brain-age model likely underestimates BAGs compared to a model trained on healthy controls only and may thus be less sensitive regarding the detection of accelerated brain-aging. This is caused by the various different disease pictures already contained in the model. However, the normative sample used here may already contain subjects who show disease specific brain alterations which might then erroneously be interpreted as biological aging during model training.

In spite of this potential lower sensitivity, the individual probabilities for accelerated brain-aging are associated with mental health diagnoses, depressive symptom severity, and body-mass index indicating a potential of individual brain-age acceleration probabilities as general risk markers. To maximize sensitivity, future studies ought to train our model on healthy individuals free from comorbidities.

Considering specificity, the fact that our model is based on a normative sample of the German population implies that higher probabilities for accelerated brain-aging might indicate especially strong deviation from neurotypicality. To investigate this, future studies ought to focus on patient groups with known deviations of brain morphology including e.g. multiple sclerosis, AD or other neurodegenerative diseases.

To assess clinical utility, prospective screening studies will be necessary which can directly quantify disease risk associated with increased individual brain-age acceleration for different diseases and known risk factors.

In order to lower the technical hurdle of application of uncertainty estimation and conformal predictions, there is a need for easily accessible software implementing the algorithms. While uncertainty estimation is crucial to virtually all clinical applications, the vast majority of studies is currently not employing uncertainty estimation with the required statistical guarantees. However, this is essential for building an AI ecosystem in medicine[24] and thereby empowering the translation from group analysis to personalized medicine. To provide easy access for the community, we implemented conformal predictions in photonai[19] based on the MAPIE toolbox[25] available at [26].

# Acknowledgments

**NAKO Investigators:**
Jochen G. Hirsch
Fraunhofer MEVIS, Bremen, Germany.

Thoralf Niendorf, Beate Endemann
Berlin Ultrahigh Field Facility (B.U.F.F.), NAKO imaging site Berlin, Max-Delbrueck Center for Molecular Medicine in the Helmholtz Association, Berlin, Germany.

Fabian Bamberg
Department of Radiology, Medical Center–University of Freiburg, Faculty of Medicine, University of Freiburg, Freiburg, Germany.

Thomas Kröncke
Department of Diagnostic and Interventional Radiology, University Hospital Augsburg, Augsburg, Germany.
Centre for Advanced Analytics and Predictive Sciences (CAAPS), Augsburg University, Augsburg, Germany

Robin Bülow
Institute of Diagnostic Radiology and Neuroradiology, University of Greifswald, Greifswald, Germany.

Henry Völzke
Institute for Community Medicine, University Medicine Greifswald, Greifswald, Germany.

Oyunbileg von Stackelberg, Ramona Felizitas Sowade
Department of Diagnostic and Interventional Radiology, University Hospital Heidelberg, Heidelberg, Germany.
Translational Lung Research Center, Member of the German Lung Research Center, Heidelberg, Germany.

Lale Umutlu, Börge Schmidt
Institute for Medical Informatics, Biometry and Epidemiology, University of Duisburg-Essen, Duisburg, Germany.

Svenja Caspers
Institute for Anatomy I, Medical Faculty and University Hospital Düsseldorf, Heinrich Heine University Düsseldorf, 40225 Düsseldorf, Germany.
Institute of Neuroscience and Medicine (INM-1), Research Centre Jülich, 52425 Jülich, Germany.

**Funding:** This work was funded by the German Research Foundation (DFG grants HA7070/2-2, HA7070/3, and HA7070/4 to T.H.) and the Interdisciplinary Center for Clinical Research (IZKF) of the medical faculty of Münster (grants Dan3/012/17 to U.D. and MzH 3/020/20 to T.H.).

The analysis was conducted with data from the GNC (www.nako.de). The GNC is funded by the Federal Ministry of Education and Research (BMBF) (project funding reference numbers: 01ER1301A/B/C and 01ER1511D), the federal states, and the Helmholtz Association with additional financial support by the participating universities and the institutes of the Helmholtz Association and of the Leibniz Association. We thank all participants who took part in the GNC study and the staff in this research program.

The BiDirect Study is supported by grants from the German Ministry of Research and Education (BMBF) to the University of Muenster (grant numbers 01ER0816 and 01ER1506).

The MACS dataset used in this work is part of the German multicenter consortium "Neurobiology of Affective Disorders: A translational perspective on brain structure and function," funded by the German Research Foundation (Deutsche Forschungsgemeinschaft DFG; Forschungsgruppe/Research Unit FOR2107). Principal investigators (PIs) with respective areas of responsibility in the FOR2107 consortium are as follows: Work Package WP1, FOR2107/MACS cohort and brain imaging: T. Kircher (speaker FOR2107; DFG grant numbers KI 588/14-1 and KI 588/14-2), U. Dannlowski (co-speaker FOR2107; DA 1151/5-1 and DA 1151/5-2), Axel Krug (KR 3822/5-1 and KR 3822/7-2), I. Nenadic (NE 2254/1-2), and C. Konrad (KO 4291/3-1). WP2, animal phenotyping: M. Wöhr (WO 1732/4-1 and WO 1732/4-2) and R. Schwarting (SCHW 559/14-1 and SCHW 559/14-2). WP3, miRNA: G. Schratt (SCHR 1136/3-1 and 1136/3-2). WP4, immunology, mitochondriae: J. Alferink (AL 1145/5-2), C. Culmsee (CU 43/9-1 and CU 43/9-2), and H. Garn (GA 545/5-1 and GA 545/7-2). WP5, genetics: M. Rietschel (RI 908/11-1 and RI 908/11-2), M. Nöthen (NO 246/10-1 and NO 246/10-2), and S. Witt (WI 3439/3-1 and WI 3439/3-2). WP6, multimethod data analytics: A. Jansen (JA 1890/7-1 and JA 1890/7-2), T. Hahn (HA 7070/2-2), B. Müller-Myhsok (MU1315/8-2), and A. Dempfle (DE 1614/3-1 and DE 1614/3-2). CP1, biobank: P. Pfefferle (PF 784/1-1 and PF 784/1-2) and H. Renz (RE 737/20-1 and 737/20-2). CP2, administration: T. Kircher (KI 588/15-1 and KI 588/17-1), U. Dannlowski (DA 1151/6-1), and C. Konrad (KO 4291/4-1). Data access and responsibility: All PIs take responsibility for the integrity of the respective study data and their components. All authors and coauthors had full access to all study data. The FOR2107 cohort project (WP1) was approved by the Ethics Committees of the Medical Faculties, University of Marburg (AZ: 07/14) and University of Münster (AZ: 2014-422-b-S).

Data collection and sharing for this project was funded by the Alzheimer's Disease Neuroimaging Initiative (ADNI) (National Institutes of Health Grant U01 AG024904) and DOD ADNI (Department of Defense award number W81XWH-12-2-0012). ADNI is funded by the National Institute on Aging, the National Institute of Biomedical Imaging and Bioengineering, and through generous contributions from the following: AbbVie, Alzheimer's Association; Alzheimer's Drug Discovery Foundation; Araclon Biotech; BioClinica, Inc.; Biogen; Bristol-Myers Squibb Company; CereSpir, Inc.; Cogstate; Eisai Inc.; Elan Pharmaceuticals, Inc.; Eli Lilly and Company; EuroImmun; F. Hoffmann-La Roche Ltd and its affiliated company Genentech, Inc.; Fujirebio; GE Healthcare; IXICO Ltd.; Janssen Alzheimer Immunotherapy Research & Development, LLC.; Johnson & Johnson Pharmaceutical Research & Development LLC.; Lumosity; Lundbeck; Merck & Co., Inc.; Meso Scale Diagnostics, LLC.; NeuroRx Research; Neurotrack Technologies; Novartis Pharmaceuticals Corporation; Pfizer Inc.; Piramal Imaging; Servier; Takeda Pharmaceutical Company; and Transition Therapeutics. The Canadian Institutes of Health Research is providing funds to support ADNI clinical sites in Canada. Private sector contributions are facilitated by the Foundation for the National Institutes of Health (www.fnih.org). The grantee organization is the Northern California Institute for Research and Education, and the study is coordinated by the Alzheimer's Therapeutic Research Institute at the University of Southern California. ADNI data are disseminated by the Laboratory for Neuro Imaging at the University of Southern California.

We thank all study participants and staff at the GNC study centers, the data management center, and the GNC head office who enabled the conduction of the study and made the collection of all data possible.

# Supplementary Material

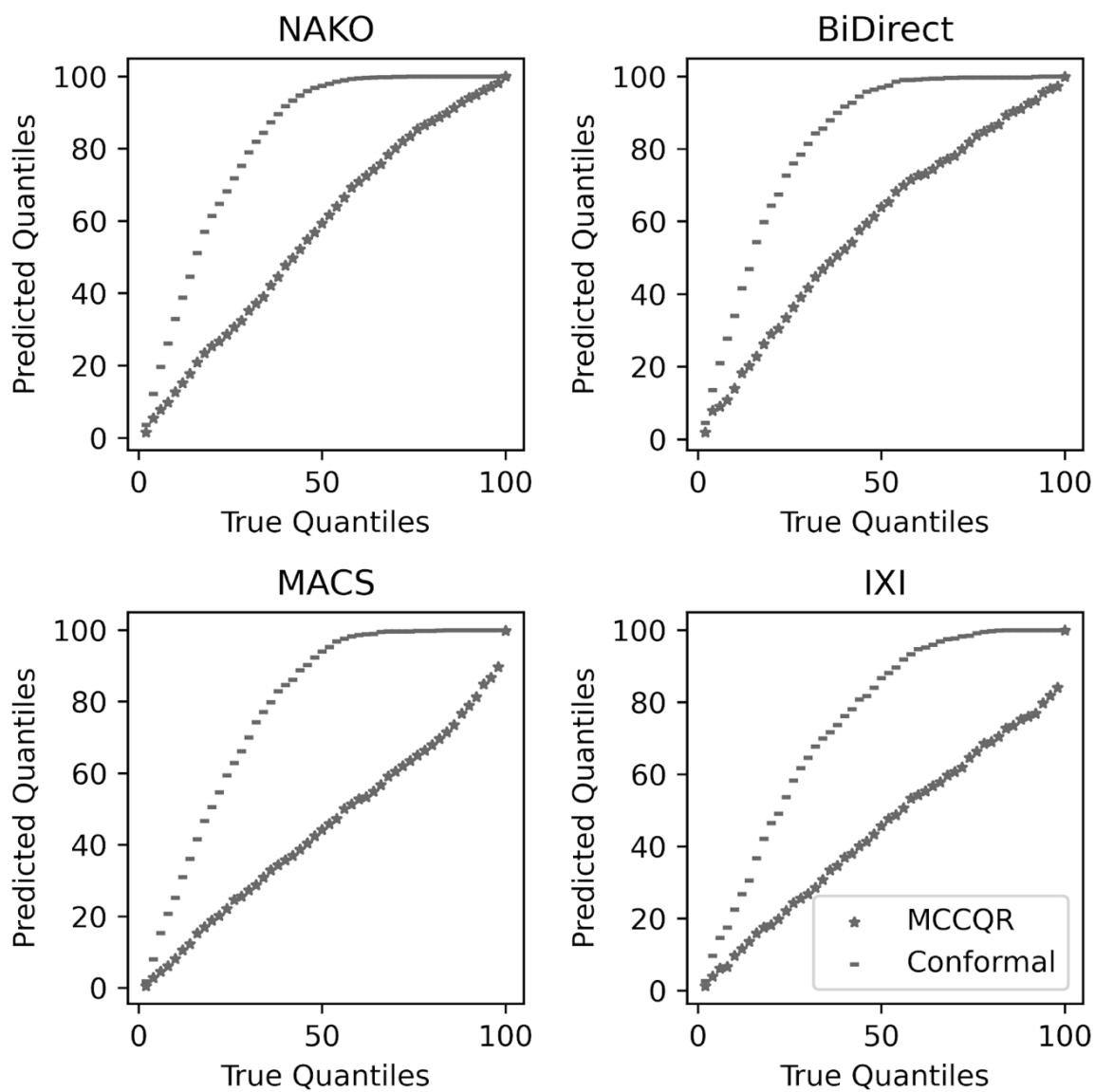

**Supplementary Figure S1.** Prediction Interval Coverage Probabilities (PICP) for 10-fold-cv NAKO and independent validation samples (BiDirect, MACS, IXI) comparing baseline MCCQR model and conformal predictions enhanced model.

| Sample | Group | N | N Males | Age Mean | Age Std. | Age Min. | Age Max. |
|---|---|---|---|---|---|---|---|
| **NAKO** | **Full Sample** | **10,691** | **5,485** | **51.79** | **11.37** | **20.00** | **72.00** |
| **BiDirect** | **Full Sample** | **1,460** | **649** | **51.37** | **8.15** | **30.91** | **70.25** |
| | MDD | 719 | 281 | 49.39 | 7.40 | 30.91 | 67.09 |
| | Cardiac Event | 53 | 12 | 56.96 | 5.97 | 43.18 | 66.54 |
| | Population Sample | 688 | 327 | 53.01 | 8.48 | 35.19 | 70.25 |
| **MACS** | **Full Sample** | **1,877** | **674** | **35.87** | **13.03** | **18.00** | **65.00** |
| | HC | 924 | 329 | 34.16 | 12.88 | 18.00 | 65.00 |
| | MDD | 822 | 285 | 36.51 | 13.18 | 18.00 | 65.00 |
| | BD | 131 | 60 | 41.76 | 11.73 | 20.00 | 64.00 |
| **IXI** | **HC** | **561** | **311** | **48.62** | **16.49** | **19.98** | **86.32** |
| **ADNI** | **Full Sample** | **2205** | **1169** | **73.21** | **7.22** | **54.4** | **91.4** |
| | HC | 799 | 354 | 72.74 | 6.32 | 55.8 | 90.1 |
| | MCI | 1014 | 593 | 72.94 | 7.56 | 54.4 | 91.4 |
| | AD | 392 | 222 | 74.88 | 7.78 | 55.1 | 90.9 |

**Supplementary Table S1.** Overview of sample and subsample distributions. Std.: Standard Deviation, Min.: Minimum, Max.: Maximum, HC: Healthy Controls, MDD: Major Depressive Disorder, BD: Bipolar Disorder, MCI: Mild Cognitive Impairment, AD: Alzheimer's Disease